\documentclass{article}

\usepackage{ijcai13}

\newtheorem {definition}{Definition}

\newtheorem {theorem}{Theorem}

\newtheorem {example}{Example}

\usepackage{amsmath}
\usepackage{amssymb}
\usepackage{rotating}

\begin{document}

\title{Logical Stochastic Optimization}

\author{ Emad Saad \\
emsaad@gmail.com
}

\maketitle

\begin{abstract}
We present a logical framework to represent and reason about stochastic optimization problems based on probability answer set programming \cite{Saad_NHPP,Saad_EHPP,Saad_DHPP}. This is established by allowing probability optimization aggregates, e.g., minimum and maximum in the language of probability answer set programming to allow minimization or maximization of some desired criteria under the probabilistic environments. We show the application of the proposed logical stochastic optimization framework under the probability answer set programming to two stages stochastic optimization problems with recourse.

\end{abstract}

\section{Introduction}

Probability answer set programming \cite{Saad_NHPP,Saad_EHPP,Saad_DHPP} is a declarative programming framework which aims to solve hard search problems in probability environments, and shown effective for probability knowledge representation and probability reasoning applications. It has been shown that many interesting probability reasoning problems are represented and solved by probability answer set programming, where probability answer sets describe the set of possible solutions to the problem. These probability reasoning problems include, but not limited to, reasoning about actions with probability effects and probability planning \cite{SaadPlan}, reinforcement learning in MDP environments \cite{Saad_MDP}, reinforcement learning in POMDP environments \cite{Saad_Learn_Sense}, contingent probability planning \cite{Saad_Sensing}, and Bayesian reasoning \cite{SaadSSAT}. However, the unavailability of probability optimization aggregates, e.g., minimum and maximum in the language of probability answer set programming \cite{Saad_NHPP,Saad_EHPP,Saad_DHPP} disallows the natural and concise representation of many interesting stochastic optimization problems that are based on minimization and maximization of some desired criteria imposed by the problem. The following stochastic optimization with recourse problem illuminates the need for these aggregates.

\begin{example} Assume that a company produces some product, $G$, and need to make a decision on the amount of units of $G$ to produce based on the market demand. The company made a decision on the amounts of units of product $G$ to produce at cost of \$$2$ per unit of $G$ (first stage). However, market demand is stochastic with a discrete probability distribution and the market demand must be met in any scenario. The company can produce extra units of product $G$ to meet the market observed demands but with the cost of \$$3$ per unit (second stage). This means a recourse to extra production to meet the excess in demand. Assume that the probability distribution, $p_i$, over market demand, $D_i$, is given as follows where two scenarios are available, $D_1 = 500$ with $p_1 = 0.6$ and $D_2 = 700$ with $p_2 = 0.4$. Formally, let $x$ be the number of units of product $G$ the company produces at the first stage and let $y_i$, called recourse variable, be the number of units the company produces at the second stage to meet the market stochastic demand at scenario $i$. The objective is to {\em minimize the total expected cost}. This two stages stochastic optimization problem is formalized as:
\[
minimize \; 2x + \sum_{i=1}^I p_i (3y_i)
\]
subject to
\[
\begin{array}{lcl}
x + y_i \geq D_i   &&    i = 1, \ldots, I \\
x \geq  0          &&                   \\
y_i \geq 0          && i = 1, \ldots, I
\end{array}
\]
where the constraint $x + y_i \geq D_i$ guarantee that demand is always met in any scenario and $I = 2$. The optimal solution to this two stages stochastic optimization with recourse problem is $x = 500$, $y_1 = 0$, $y_2 = 200$, and with minimum total expected cost equal to $\$1240$.
\label{ex:finance}
\end{example}
To represent this stochastic optimization problem in probability answer set programming and to provide correct solution to the problem, the probability answer set programming representation of the problem has to be able to represent the probability distributions of the problem domain and any probability distribution that may arise to the problem constraints along with the preference relation that minimizes or maximizes the objective function including the expected values that always appear in the objective functions of these types of stochastic optimization problems, and to be able to compare for the minimum or the maximum of the objective value across the generated probability answer sets.

However, the current syntax and semantics of probability answer set programming do not define probability preference relations or rank probability answer sets based on minimization or maximization of some desired criterion specified by the user. Therefore, in this paper we extend probability answer set programming with probability aggregate preferences to allow the ability to represent and reason and intuitively solve stochastic optimization problems. The proposed probability aggregates probability answer set optimization framework presented in this paper modifies and generalizes the classical aggregates classical answer set optimization presented in \cite{Saad_ASOG} as well as the classical answer set optimization introduced in \cite{ASO}. We show the application of probability aggregates probability answer set optimization to a two stages stochastic optimization with recourse problem described in Example (\ref{ex:finance}), where a probability answer set program \cite{Saad_DHPP} (disjunctive hybrid probability logic program with probability answer set semantics) is used as probability answer sets generator rules.

\section{Probability Aggregates Probability Answer Set Optimization}

Probability answer set optimization programs are probability logic programs under the probability answer set semantics whose probability answer sets are ranked according to probability preference rules represented in the programs. A probability answer set optimization program, $\Pi$, is a pair of the form \\ $\Pi = \langle R_{gen} \cup R_{pref}, \tau \rangle$, where $R_{gen} \cup R_{pref}$ is a union of two sets of probability logic rules and $\tau$ is a mapping, $\tau: {\cal B_L} \rightarrow S_{disj}$, associated to the set of probability logic rules $R_{gen}$. The first set of probability logic rules, $R_{gen}$, is called the generator rules that generate the probability answer sets that satisfy every probability logic rule in $R_{gen}$ and the mapping $\tau$ associates to each atom, $a$, appearing in $R_{gen}$, a disjunctive p-strategy that is used to combine the probability intervals obtained from different probability logic rules in $R_{gen}$ with an atom $a$ appearing in their heads. $R_{gen}$ is any set of probability logic rules with well-defined probability answer set semantics including normal, extended, and disjunctive hybrid probability logic rules \cite{Saad_NHPP,Saad_EHPP,Saad_DHPP}, as well as hybrid probability logic rules with probability aggregates (all are forms of {\em probability answer set programming}).

The second set of probability logic rules, $R_{pref}$, is called the {\em probability preference rules}, which are probability logic rules that represent the user's {\em probability quantitative} and {\em qualitative preferences} over the probability answer sets generated by $R_{gen}$. The probability preference rules in $R_{pref}$ are used to rank the generated probability answer sets from $R_{gen}$ from the top preferred probability answer set to the least preferred probability answer set. Similar to \cite{ASO}, an advantage of probability answer set optimization programs is that $R_{gen}$ and $R_{pref}$ are independent. This makes probability preference elicitation easier and the whole approach is more intuitive and easy to use in practice.

In our introduction of probability answer set optimization programs, we focus on the syntax and semantics of the {\em probability preference rules}, $R_{pref}$, of the probability answer set optimization programs, since the syntax and semantics of the probability answer sets generator rules, $R_{gen}$, are the same as syntax and semantics of any set of probability logic rules with well-defined probability answer set semantics as described in \cite{Saad_NHPP,Saad_EHPP,Saad_DHPP}.


\subsection{Basic Language}

Let ${\cal L}$ be a first-order language with finitely many predicate symbols, function symbols, constants, and
infinitely many variables. A literal is either an atom $a$ in ${\cal B_L}$ or the negation of an atom $a$ ($\neg a$), where ${\cal B_L}$ is the Herbrand base of ${\cal L}$ and $\neg$ is the classical negation. Non-monotonic negation or the negation as failure is denoted by $not$. The Herbrand universe of $\cal L$ is denoted by $U_{\cal L}$. Let $Lit$ be the set of all literals in ${\cal L}$, where $Lit = \{a \:| \: a \in {\cal B_L} \} \cup \{\neg a \: | \: a \in {\cal B_L}\}$. A \emph{probability annotation} is a probability interval of the form $[\alpha_1, \alpha_2]$, where $\alpha_1, \alpha_2$ are called probability annotation items. A \emph{probability annotation item} is either a constant in $[0, 1]$ (called {\em probability annotation constant}), a variable ranging over $[0, 1]$ (called \emph{probability annotation variable}), or $f(\alpha_1,\ldots,\alpha_n)$ (called \emph{probability annotation function}) where $f$ is a representation of a monotone, antimonotone, or nonmonotone total or partial function $f: ([0, 1])^n \rightarrow [0, 1]$ and $\alpha_1,\ldots,\alpha_n$ are probability annotation items.

Let $S = S_{conj} {\cup} S_{disj}$ be an arbitrary set of p-strategies, where $S_{conj}$ ($S_{disj}$) is the set of all conjunctive (disjunctive) p-strategies in $S$. A \emph{hybrid literals} is an expression of the form  $l_1 \wedge_\rho \ldots \wedge_\rho l_n$ or $l_1 \vee_{\rho'} \ldots \vee_{\rho'} l_n$, where $l_1, \ldots, l_n$ are literals and $\rho$ and $\rho'$ are p-strategies from $S$. $bf_S(Lit)$ is the set of all ground hybrid literals formed using distinct literals from $Lit$ and p-strategies from $S$. If $L$ is a hybrid literal $\mu$ is a probability annotation then $L:\mu$ is called a probability annotated hybrid literal.

A symbolic probability set is an expression of the form $\{ X : [P_1, P_2] \; | \; C \}$, where $X$ is a variable or a function term and $P_1$, $P_2$ are probability annotation variables or probability annotation functions, and $C$ is a conjunction of probability annotated hybrid basic formulae. A ground probability set is a set of pairs of the form $\langle x : [p_1, p_2] \; | \; C^g \rangle$ such that $x$ is a constant term and $p_1, p_2$ are probability annotation constants, and $C^g$ is a ground conjunction of probability annotated hybrid basic formulae. A symbolic probability set or ground probability set is called a probability set term. Let $f$ be a probability aggregate function symbol and $S$ be a probability set term, then $f(S)$ is said a probability aggregate, where $f \in \{$ $val_E$, $sum_E$, $times_E$, $min_E$, $max_E$, $count_E$, $sum_P$, $times_P$, $min_P$, $max_P$, $count_P$  $\}$. If $f(S)$ is a probability aggregate and $T$ is an interval $[\theta_1, \theta_2]$, called {\em guard}, where $\theta_1, \theta_2$ are constants, variables or functions terms, then we say $f(S) \prec T$ is a probability aggregate atom, where $\prec \in \{=, \neq, <, >, \leq, \geq \}$.

A {\em probability optimization aggregate} is an expression of the form $max_\mu (f(S))$, $min_\mu (f(S))$, $max_x (f(S))$, $min_x (f(S))$, $max_{x \mu} (f(S))$, and $min_{x \mu} (f(S))$, where $f$ is a probability aggregate function symbol and $S$ is a probability set term.

\subsection{Probability Preference Rules Syntax}

Let ${\cal A}$ be a set of probability annotated hybrid literals, probability annotated probability aggregate atoms and probability optimization aggregates. A boolean combination over ${\cal A}$ is a boolean formula over probability annotated hybrid literals, probability annotated probability aggregate atoms, and probability optimization aggregates in ${\cal A}$ constructed by conjunction, disjunction, and non-monotonic negation ($not$), where non-monotonic negation is combined only with probability annotated hybrid literals and probability annotated probability aggregate atoms

\begin{definition} A probability preference rule, $r$, over a set of probability annotated hybrid literals, probability annotated probability aggregate atoms and probability optimization aggregates, ${\cal A}$, is an expression of the form
\begin{eqnarray}
C_1 \succ  C_2 \succ \ldots \succ C_k \leftarrow L_{k+1}:\mu_{k+1},\ldots, L_m:\mu_m, \notag \\
not\; L_{m+1}:\mu_{m+1},\ldots, not\;L_{n}:\mu_{n}
\label{rule:pref}
\end{eqnarray}
where $L_{k+1}:\mu_{k+1}, \ldots, L_{n}:\mu_{n}$ are probability annotated hybrid literals and probability annotated probability aggregate atoms and $C_1, C_2, \ldots, C_k$ are boolean combinations over ${\cal A}$.
\end{definition}
Let $body(r) = L_{k+1}:\mu_{k+1},\ldots, L_m:\mu_m, not\; L_{m+1}:\mu_{m+1},\ldots, not\;L_{n}:\mu_{n}$ and $head(r) = C_1 \succ C_2 \succ \ldots \succ C_k$, where $r$ is a probability preference rule of the form (\ref{rule:pref}). Intuitively, a probability preference rule, $r$, of the form (\ref{rule:pref}) means that any probability answer set that satisfies $body(r)$ and $C_1$ is preferred over the probability answer sets that satisfy $body(r)$, some $C_i$ $(2 \leq i \leq k)$, but not $C_1$, and any probability answer set that satisfies $body(r)$ and $C_2$ is preferred over probability answer sets that satisfy $body(r)$, some $C_i$ $(3 \leq i \leq k)$, but neither $C_1$ nor $C_2$, etc.

\begin{definition} A probability answer set optimization program, $\Pi$, is a pair of the form $\Pi = \langle R_{gen} \cup R_{pref}, \tau \rangle$, where $R_{gen}$ is a set of probability logic rules with well-defined probability answer set semantics, the {\em generator} rules, $R_{pref}$ is a set of probability preference rules, and $\tau$ is the mapping $\tau: {\cal L}it \rightarrow S_{disj}$ that associates to each literal, $l$, appearing in $R_{gen}$ a disjunctive p-strategy.
\end{definition}
Let $f(S)$ be a probability aggregate. A variable, $X$, is a local variable to $f(S)$ if and only if $X$ appears in $S$ and $X$ does not appear in the probability preference rule that contains $f(S)$. A global variable is a variable that is not a local variable. Therefore, the {\em ground instantiation} of a symbolic probability set $$S = \{ X:[P_1,P_2]  \; | \; C \}$$ is the set of all ground pairs of the form $\langle \theta\; (X) :[\theta\; (P_1), \theta\; (P_2)]\; | \;  \theta \; (C) \rangle$, where $\theta$ is a substitution of every local variable appearing in $S$ to a constant from $U_{\cal L}$. A ground instantiation of a probability preference rule, $r$, is the replacement of each global variable appearing in $r$ to a constant from $U_{\cal L}$, then followed by the ground instantiation of every symbolic probability set, $S$, appearing in $r$. The ground instantiation of a probability aggregates probability answer set optimization program, $\Pi$, is the set of all possible ground instantiations of every probability rule in $\Pi$.


\begin{example} The two stages stochastic optimization with recourse problem presented in Example (\ref{ex:finance}) can be represented as a probability aggregates probability answer set optimization program $\Pi = \langle R_{gen} \cup R_{pref}, \tau \rangle$, where $\tau$ is any assignments of disjunctive p-strategies and $R_{gen}$ is a set of disjunctive hybrid probability logic rules with probability answer set semantics \cite{Saad_DHPP} of the form:
\[
\begin{array}{r}
domX(500) \vee  domX(550) \vee domX(600) \vee  domX(650) \vee \\  domX(700). \\
domY_1(0):p_1  \vee domY_1(50):p_1   \vee  domY_1(100):p_1  \vee   \\ domY_1(150):p_1 \vee domY_1(200):p_1.  \\
domY_2(0):p_2 \vee domY_2(50):p_2  \vee  domY_2(100):p_2 \vee \\ domY_2(150):p_2  \vee  domY_2(200):p_2.
\end{array}
\]
\[
\begin{array}{r}
objective(X,Y_1,Y_2, 2 * X + 3 * p_1 * Y_1 + 3 * p_2 * Y_2)  \\ \leftarrow  domX(X), domY_1(Y_1):p_1, \\   domY_1(Y_2):p_2.
\end{array}
\]
\[
\begin{array}{lcl}
& \leftarrow & domX(X), domY_1(Y_1):p_1, X + Y_1 < 500.\\
& \leftarrow & domX(X), domY_2(Y_2):p_2, X + Y_2 < 700.
\end{array}
\]
where $p_1 = 0.6$ and $p_2 = 0.4$, $domX(X)$, $domY_1(Y_1)$, $domY_2(Y_2)$ are predicates represent the domains of possible values for the variables $X$, $Y_1$, $Y_2$ that represent the units of product $G$ corresponding to the variables $x, y_1, y_2$ described in Example (\ref{ex:finance}), $objective(X, Y_1, Y_2, Cost)$ is a predicate that represents the objective value, $Cost$, for the assignments of units of a product $G$ to the variables $X$, $Y_1$, $Y_2$ where $Cost$ is the expected cost for this assignment of variables.

The set of probability preference rules, $R_{pref}$, of $\Pi$ consists of the probability preference rule
\[
\begin{array}{r}
min_x \{Cost : 1 \; | \; objective(X,Y_1,Y_2, Cost) \} \leftarrow
\end{array}
\]
\label{ex:finance-code}
\end{example}

\section{Probability Aggregates Probability Answer Set Optimization Semantics}

Let $\mathbb{X}$ denotes a set of objects. Then, we use $2^\mathbb{X}$ to denote the set of all {\em multisets} over elements in $\mathbb{X}$. Let $C[0, 1]$ denotes the set of all closed intervals in $[0, 1]$, $\mathbb{R}$ denotes the set of all real numbers, $\mathbb{N}$ denotes the set of all natural numbers, and $U_{\cal L}$ denotes the Herbrand universe. Let $\bot$ be a symbol that does not occur in ${\cal L}$. Therefore, the semantics of the probability aggregates are defined by the mappings:

\begin{itemize}

\item $val_E :  2^{\mathbb{R} \times {\cal C}[0, 1]} \rightarrow [\mathbb{R}, \mathbb{R}]$.

\item $sum_E :  2^{\mathbb{R} \times {\cal C}[0, 1] } \rightarrow [\mathbb{R}, \mathbb{R}]$.

\item $times_E: 2^{\mathbb{R} \times {\cal C}[0, 1] } \rightarrow [\mathbb{R}, \mathbb{R}]$.

\item $min_E, max_E: (2^{\mathbb{R} \times {\cal C}[0, 1]} - \emptyset) \rightarrow [\mathbb{R}, \mathbb{R}]$.

\item $count_E : 2^{U_{\cal L} \times {\cal C}[0, 1]} \rightarrow [\mathbb{R}, \mathbb{R}]$.

\item $sum_P : 2^{\mathbb{R} \times {\cal C}[0, 1] } \rightarrow \mathbb{R} \times {\cal C}[0, 1]$.

\item $times_P: 2^{\mathbb{R} \times {\cal C}[0, 1] } \rightarrow \mathbb{R} \times {\cal C}[0, 1]$.

\item $min_P, max_P: (2^{\mathbb{R} \times {\cal C}[0, 1] } - \emptyset) \rightarrow
    \mathbb{R} \times {\cal C}[0, 1]$.

\item $count_P : 2^{U_{\cal L} \times {\cal C}[0, 1]}  \rightarrow \mathbb{N} \times {\cal C}[0, 1]$.

\end{itemize}
The application of $sum_E$ and $times_E$ on the empty multiset return $[0,0]$ and $[1,1]$ respectively. The application of $val_E$ and $count_E$ on the empty multiset returns $[0,0]$. The application of $sum_P$ and $times_P$ on the empty multiset return $(0,[1,1])$ and $(1,[1,1])$ respectively. The application of $count_P$ on the empty multiset returns $(0,[1,1])$. However, the application of $max_E$, $min_E$, $max_P$, $min_P$ on the empty multiset is undefined.

The semantics of probability aggregates and probability optimization aggregates in probability aggregates probability answer set optimization is defined with respect to a probability answer set, which is, in general, a total or partial mapping, $h$, from $bf_S({\cal L}it)$ to ${\cal C}[0,1]$. In addition, the semantics of probability optimization aggregates $max_\mu (f(S))$, $min_\mu (f(S))$, $max_x (f(S))$, $min_x (f(S))$, $max_{x \mu} (f(S))$, and $min_{x \mu} (f(S))$ are based on the semantics of the probability aggregates $f(S)$.

We say, a probability annotated hybrid literal, $L\mu$, is true (satisfied) with respect to a probability answer set, $h$, if and only if $\mu \leq h(L)$. The negation of a probability hybrid literal, $not \; L:\mu$, is true (satisfied) with respect to $h$ if and only if $\mu \nleq h(L)$ or $L$ is undefined in $h$. The evaluation of probability aggregates and the truth valuation of probability aggregate atoms with respect to probability answer sets are given as follows. Let $f(S)$ be a ground probability aggregate and $h$ be a probability answer set. In addition, let $S_h$ be the multiset constructed from elements in $S$, where $S_h = \{\!\!\{ x : [p_1,p_2]  \; | \; \langle x : [p_1,p_2] \; | \; C^g \rangle \in S \wedge$ $C^g$ is true w.r.t. $h \}\!\!\}$. Then, the evaluation of $f(S)$ with respect to $h$ is, $f(S_h)$, the result of the application of $f$ to $S_h$, where $f(S_h) = \bot$ if $S_h$ is not in the domain of $f$ and

\begin{itemize}

\item $val_E(S_h) = \sum_{x : [p_1, p_2] \in S_h} \;( x \times [p_1, p_2])$

\item $sum_E(S_h) = (\sum_{x : [p_1, p_2] \in S_h} \; x) \; \times \; (\prod_{x : [p_1, p_2] \in S_h} \; [p_1, p_2])$

\item $times_E(S_h) = (\prod_{x : [p_1, p_2] \in S_h} \; x ) \; \times  \; (\prod_{x : [p_1, p_2] \in S_h} \; [p_1, p_2])$

\item $min_E (S_h)= (\min_{x : [p_1, p_2] \in S_h} \; x ) \; \times \; (\prod_{x : [p_1, p_2] \in S_h} \; [p_1, p_2])$

\item $max_E (S_h)= (\max_{x : [p_1, p_2] \in S_h} \; x ) \; \times \; (\prod_{x : [p_1, p_2] \in S_h} \; [p_1, p_2])$

\item $count_E (S_h)= (count_{x : [p_1, p_2] \in S_h} \; x )  \; \times \; (\prod_{x : [p_1, p_2] \in S_h} \; [p_1, p_2])$
\\

\item $sum_P (S_h) = (\sum_{x : [p_1, p_2] \in S_h} \; x  \;, \; \prod_{x : [p_1, p_2] \in S_h} \; [p_1, p_2])$

\item $times_P (S_h) = (\prod_{x : [p_1, p_2] \in S_h} \; x  \;, \; \prod_{x : [p_1, p_2] \in S_h} \; [p_1, p_2])$

\item $min_P (S_h) = (\min_{x : [p_1, p_2] \in S_h} \; x \; , \; \prod_{x : [p_1, p_2] \in S_h} \; [p_1, p_2])$

\item $max_P (S_h) = (\max_{x : [p_1, p_2] \in S_h} \; x \; , \; \prod_{x : [p_1, p_2] \in S_h} \; [p_1, p_2])$

\item $count_P (S_h) = (count_{x : [p_1, p_2] \in S_h} \; x \; , \; \prod_{x : [p_1, p_2] \in S_h} \; [p_1, p_2])$

\end{itemize}

\subsection{Probability Preference Rules Semantics}

In this section, we define the notion of satisfaction of probability preference rules with respect to probability answer sets. We consider that probability annotated probability aggregate atoms that involve probability aggregates from $\{$$val_E$, $sum_E$, $times_E$, $min_E$, $max_E$, $count_E$$\}$ are associated to the probability annotation $[1,1]$.

Let  $\Pi = \langle R_{gen} \cup R_{pref}, \tau \rangle$ be a ground probability aggregates probability answer set optimization program, $h,h'$ be probability answer sets for $R_{gen}$ (possibly partial), $f \in \{$$val_E$, $sum_E$, $times_E$, $min_E$, $max_E$, $count_E$$\}$ and $g \in \{$$sum_P$, $times_P$, $min_P$, $max_P$, $count_P$$\}$, and $r$ be a probability preference rule in $R_{pref}$. Then the satisfaction of a boolean combination, $C$, appearing in $head(r)$, by $h$ is defined inductively as follows:

\begin{itemize}

\item $h$ satisfies $L:\mu$ iff  $\mu \leq h(L)$.

\item $h$ satisfies $not\;L:\mu$ iff $\mu \nleq h(L)$ or $L$ is undefined in $h$.
\\

\item $h$ satisfies $f(S) \prec T : [1,1]$ iff $f(S_h) \neq \bot$ and $f(S_h) \prec T$.

\item $h$ satisfies $not \; f(S) \prec T :[1,1] $ iff $f(S_h) =  \bot$ or  $f(S_h) \neq \bot$ and $f(S_h) \nprec T$.

\item $h$ satisfies $g(S) \prec T : \mu$  iff $g(S_h) = (x,\nu) \neq \bot$ and $x \prec T$ and $\mu \leq_t \nu$.

\item $h$ satisfies $not \; g(S) \prec T :\mu $ iff $g(S_h) =  \bot$ or $g(S_h) = (x, \nu ) \neq \bot$ and $x \nprec T$ or $\mu \nleq_t \nu$.
\\

\item $h$ satisfies $max (f(S))$ iff $f(S_h) = x \neq \bot$ and for any $h'$, $f(S_{h'}) = x' \neq \bot$ and $x' \leq x$ or $f(S_h) \neq \bot$ and $f(S_{h'}) = \bot$.

\item $h$ satisfies $min (f(S))$ iff $f(S_h) = x \neq \bot$ and for any $h'$, $f(S_{h'}) = x' \neq \bot$ and $x \leq x'$ or $f(S_h) \neq \bot$ and $f(S_{h'}) = \bot$.
\\

\item $h$ satisfies $max_\mu (g(S))$ iff $g(S_h) = (x, \nu) \neq \bot$ and for any $h'$, $g(S_{h'}) = (x', \nu') \neq \bot$ and $\nu' \leq \nu$ or $g(S_h) \neq \bot$ and $g(S_{h'}) = \bot$.

\item $h$ satisfies $min_\mu (g(S))$ iff $g(S_h) = (x, \nu) \neq \bot$ and for any $h'$, $g(S_{h'}) = (x', \nu') \neq \bot$ and $\nu \leq \nu'$ or $g(S_h) \neq \bot$ and $g(S_{h'}) = \bot$.

\item $h$ satisfies $max_x (g(S))$ iff $g(S_h) = (x, \nu) \neq \bot$ and for any $h'$, $g(S_{h'}) = (x', \nu') \neq \bot$ and $x' \leq x$ or $g(S_h) \neq \bot$ and $g(S_{h'}) = \bot$.

\item $h$ satisfies $min_x (g(S))$ iff $g(S_h) = (x, \nu) \neq \bot$ and for any $h'$, $g(S_{h'}) = (x', \nu') \neq \bot$ and $x \leq x'$ or $g(S_h) \neq \bot$ and $g(S_{h'}) = \bot$.

\item $h$ satisfies $max_{x \mu} (g(S))$ iff $g(S_h) = (x, \nu) \neq \bot$ and for any $h'$, $g(S_{h'}) = (x', \nu') \neq \bot$ and $x' \leq x$ and $\nu' \leq \nu$ or $g(S_h) \neq \bot$ and $g(S_{h'}) = \bot$.

\item $h$ satisfies $min_{x \mu} (g(S))$ iff $g(S_h) = (x, \nu) \neq \bot$ and for any $h'$, $g(S_{h'}) = (x', \nu') \neq \bot$ and $x \leq x'$ and $\nu \leq \nu'$ or $g(S_h) \neq \bot$ and $g(S_{h'}) = \bot$.
\\

\item $h \models C_1 \wedge C_2$ iff $h \models C_1$ and $h \models C_2$.

\item $h  \models C_1 \vee C_2$ iff $h \models C_1$ or $h \models C_2$.

\end{itemize}
\label{def:satisfaction}
The satisfaction of $body(r)$ by $h$ is defined inductively as follows:
\begin{itemize}

\item $h$ satisfies $L:\mu$ iff  $\mu \leq h(L)$.

\item $h$ satisfies $not\;L:\mu$ iff $\mu \nleq h(L)$ or $L$ is undefined in $h$.

\item $h$ satisfies $f(S) \prec T : [1,1]$ iff $f(S_h) \neq \bot$ and $f(S_h) \prec T$.

\item $h$ satisfies $not \; f(S) \prec T :[1,1] $ iff $f(S_h) =  \bot$ or  $f(S_h) \neq \bot$ and $f(S_h) \nprec T$.

\item $h$ satisfies $g(S) \prec T : \mu$  iff $g(S_h) = (x,\nu) \neq \bot$ and $x \prec T$ and $\mu \leq_t \nu$.

\item $h$ satisfies $not \; g(S) \prec T :\mu $ iff $g(S_h) =  \bot$ or $g(S_h) = (x, \nu ) \neq \bot$ and $x \nprec T$ or $\mu \nleq_t \nu$.

\item $h$ satisfies $body(r)$ iff $\forall(k+1 \leq i \leq m)$, $h$ satisfies $L_i : \mu_i$ and $\forall(m+1 \leq j \leq n)$, $h$ satisfies $not\; L_j : \mu_j$.
\end{itemize}
The application of the probability aggregates, $f \in \{val_E, sum_E, time_E, min_E, max_E \}$, on a singleton $\{x:\mu \}$, returns $x . \mu$ ($x$ multiplied by $\mu$), i.e., $f(\{x:\mu\}) = x . \mu$. Therefore, we use $max(S)$ and $min(S)$ as abbreviations for the probability optimization aggregates $max(f(S))$ and $min(f(S))$ respectively, whenever $S$ is a singleton and $f \in \{val_E, sum_E, time_E, min_E, max_E \}$. Similarly, the application of the probability aggregates, $g \in \{sum_P, time_P, min_P, max_P \}$, on a singleton $\{x:\mu \}$, returns $(x, \mu)$, i.e., $f(\{x:\mu\}) = (x, \mu)$. Therefore, we use $max_\mu(S)$, $min_\mu(S)$, $max_x(S)$, $min_x(S)$, $max_{x \mu}(S)$, and $min_{x \mu}(S)$ as abbreviations for the probability optimization aggregates $max_\mu(g(S))$, $min_\mu(g(S))$, $max_x(g(S))$, $min_x(g(S))$, $max_{x \mu}(g(S))$, and $min_{x \mu}(g(S))$ respectively, whenever $S$ is a singleton and $g \in \{sum_P, time_P, min_P,  max_P \}$.

\begin{definition} Let $\Pi = \langle R_{gen} \cup R_{pref}, \tau \rangle$ be a ground probability aggregates probability answer set optimization program, $h$ be a probability answer set for $R_{gen}$, and $r$ be a probability preference rule in $R_{pref}$, and $C_i$ be a boolean combination in $head(r)$. Then, we define the following notions of satisfaction of $r$ by $h$:

\begin{itemize}
\item $h \models_{i} r$ iff $h \models body(r)$ and $h \models C_i$.

\item $h \models_{irr} r$ iff $h \models body(r)$ and $h$ does not satisfy any $C_i$ in $head(r)$.

\item $h \models_{irr} r$ iff $h$ does not satisfy $body(r)$.
\end{itemize}
\end{definition}
$h \models_{i} r$ means that the body of $r$ and the boolean combination $C_i$ that appearing in the head of $r$ is satisfied by $h$. However, $h \models_{irr} r$ means that $r$ is irrelevant (denoted by $irr$) to $h$, or, in other words, the probability preference rule $r$ is not satisfied by $h$, because either one of two reasons. Either because the body of $r$ and non of the boolean combinations that appearing in the head of $r$ are satisfied by $h$. Or because the body of $r$ is not satisfied by $h$.

\subsection{Probability Answer Sets Ranking}

In this section we define the ranking of the probability answer sets with respect to a boolean combination, a probability preference rule, and with respect to a set of probability preference rules.

\begin{definition}
Let $\Pi = \langle R_{gen} \cup R_{pref}, \tau \rangle$ be a ground probability aggregates probability answer set optimization program, $h_1, h_2$ be two probability answer sets for $R_{gen}$, $r$ be a probability preference rule in $R_{pref}$, $C_i$ be boolean combination appearing in $head(r)$, and $f \in \{$$val_E$, $sum_E$, $times_E$, $min_E$, $max_E$, $count_E$$\}$ and $g \in \{$$sum_P$, $times_P$, $min_P$, $max_P$, $count_P$$\}$. Then, $h_1$ is strictly preferred over $h_2$ w.r.t. $C_i$, denoted by $h_1 \succ_i h_2$, iff $h_1 \models C_i$ and $h_2 \nvDash C_i$ or $h_1 \models C_i$ and $h_2 \models C_i$ (except $C_i$ is a probability optimization aggregate) and one of the following holds:

\begin{itemize}

\item $C_i = L:\mu$ implies $h_1 \succ_i h_2$ iff $h_1(L) > h_2(L)$.

\item $C_i = not \; L:\mu$ implies $h_1 \succ_i h_2$ iff $h_1(L) < h_2(L)$ or $L$ is undefined in $h_1$ but defined $h_2$.
\\

\item $C_i = f(S) \prec T : [1,1]$ implies $h_1 \succ_i h_2$ iff  $f(S_{h_1}) = x \neq \bot$, $f(S_{h_2}) = x' \neq \bot$, and $x' < x$.

    \item $C_i = not \; f(S) \prec T :[1,1] $ implies $h_1 \succ_i h_2$ iff

        \begin{itemize}
            \item $f(S_{h_1}) =  \bot$ and $f(S_{h_2}) \neq  \bot$ or
            \item $f(S_{h_1}) = x \neq \bot$, $f(S_{h_2}) = x' \neq \bot$, and $x < x'$
            \\

        \end{itemize}

\item $C_i = g(S) \prec T : \mu$ implies $h_1 \succ_i h_2$ iff  $g(S_{h_1}) = (x, \nu) \neq \bot$, $g(S_{h_2}) = (x', \nu') \neq \bot$, and $\nu' < \nu$.

    \item $C_i = not \; g(S) \prec T :\mu $ implies $h_1 \succ_i h_2$ iff

        \begin{itemize}
            \item $g(S_{h_1}) =  \bot$ and $g(S_{h_2}) \neq  \bot$ or
            \item $g(S_{h_1}) = (x, \nu) \neq \bot$, $g(S_{h_2}) = (x', \nu') \neq \bot$, and $\nu < \nu'$ \\

        \end{itemize}

\item $C_i \in \{ max (f(S)), \; min (f(S)), max_\mu (g(S)), \; min_\mu (g(S)), \\ max_x (g(S)), min_x (g(S)),
          max_{x \mu} (g(S)), \; min_{x \mu} (g(S)) \}$ implies $h_1 \succ_i h_2$ iff  $h_1 \models C_i$ and $h_2 \nvDash C_i$.
\\

\item $C_i = C_{i_1} \wedge C_{i_2}$ implies $h_1 \succ_i h_2$ iff there exists $t \in \{{i_1}, {i_2}\}$ such that $h_1 \succ_t h_2$ and for all other $t' \in \{{i_1}, {i_2}\}$, we have $h_1 \succeq_{t'} h_2$.

\item $C_i = C_{i_1} \vee C_{i_2}$ implies $h_1 \succ_i h_2$ iff there exists $t \in \{{i_1}, {i_2}\}$ such that $h_1 \succ_t h_2$ and for all other $t' \in \{{i_1}, {i_2}\}$, we have $h_1 \succeq_{t'} h_2$.

\end{itemize}
We say, $h_1$ and $h_2$ are equally preferred w.r.t. $C_i$, denoted by $h_1 =_{i} h_2$, iff $h_1 \nvDash C_i$ and $h_2 \nvDash C_i$ or $h_1 \models C_i$ and $h_2 \models C_i$ and one of the following holds:

\begin{itemize}

\item $C_i = L:\mu$ implies $h_1 =_{i} h_2$ iff $h_1(L) = h_2(L)$.

\item $C_i = not \; L:\mu$ implies $h_1 =_{i} h_2$  iff $h_1(L) = h_2(L)$ or $L$ is undefined in both $h_1$ and $h_2$.
 \\

 \item $C_i = f(S) \prec T : [1,1]$ implies $h_1 =_i h_2$ iff  $f(S_{h_1}) = x \neq \bot$, $f(S_{h_2}) = x' \neq \bot$, and $x' = x$.

    \item $C_i = not \; f(S) \prec T :[1,1] $ implies $h_1 =_i h_2$ iff

        \begin{itemize}
            \item $f(S_{h_1}) = \bot$ and $f(S_{h_2}) = \bot$ or
            \item $f(S_{h_1}) = f(S_{h_2})  \neq \bot$
            \\

        \end{itemize}

\item $C_i = g(S) \prec T : \mu$ implies $h_1 =_i h_2$ iff  $g(S_{h_1}) = (x, \nu) \neq \bot$, $g(S_{h_2}) = (x', \nu') \neq \bot$, and $\nu' = \nu$.

    \item $C_i = not \; g(S) \prec T :\mu $ implies $h_1 =_i h_2$ iff

        \begin{itemize}
            \item $g(S_{h_1}) = \bot$ and $g(S_{h_2}) =  \bot$ or
            \item $g(S_{h_1}) = (x, \nu) \neq \bot$, $g(S_{h_2}) = (x', \nu') \neq \bot$, and $\nu = \nu'$ \\

        \end{itemize}

\item $C_i \in \{ max (f(S)), \; min (f(S)), max_\mu (g(S)), \\ min_\mu (g(S)), max_x (g(S)), min_x (g(S)),
          max_{x \mu} (g(S)), \\ min_{x \mu} (g(S)) \}$ implies $h_1 =_i h_2$ iff  $h_1 \models C_i$ and $h_2 \models C_i$.
\\

\item $C_i = C_{i_1} \wedge C_{i_2}$ implies $h_1 =_{i} h_2$ iff
\[
\forall \: t \in \{{i_1}, {i_2}\}, \; h_1 =_{t} h_2
\]

\item $C_i = C_{i_1} \vee C_{i_2}$ implies $h_1 =_{i} h_2$ iff
\[
|\{h_1 \succeq_{t} h_2 \: | \: \forall \: t \in \{{i_1}, {i_2}\} \}| = | \{ h_2 \succeq_{t} h_1 \: | \: \forall \: t \in \{{i_1}, {i_2}\} \}|.
\]

\end{itemize}
We say, $h_1$ is at least as preferred as $h_2$ w.r.t. $C_i$, denoted by $h_1 \succeq_i h_2$, iff $h_1 \succ_i h_2$ or $h_1 =_i h_2$.
\label{def:compination}
\end{definition}

\begin{definition} Let $\Pi = \langle R_{gen} \cup R_{pref}, \tau \rangle$ be a ground probability aggregates probability answer set optimization program, $h_1, h_2$ be two probability answer sets for $R_{gen}$, $r$ be a probability preference rule in $R_{pref}$, and $C_l$ be a boolean combination appearing in $head(r)$. Then, $h_1$ is strictly preferred over $h_2$ w.r.t. $r$, denoted by $h_1 \succ_r h_2$, iff one of the following holds:
\begin{itemize}
\item $h_1 \models_{i} r$ and $h_2 \models_{j} r$ and $i < j$, \\
where $i = \min \{l \; | \; h_1 \models_l r \}$ and $j = \min \{l \; | \; h_2 \models_l r \}$.

\item $h_1 \models_{i} r$ and $h_2 \models_{i} r$ and $h_1 \succ_i h_2$, \\
where $i = \min \{l \; | \; h_1 \models_l r \} = \min \{l \; | \; h_2 \models_l r \}$.

\item $h_1 \models_{i} r$ and $h_2 \models_{irr} r$.
\end{itemize}
We say, $h_1$ and $h_2$ are equally preferred w.r.t. $r$, denoted by $h_1 =_{r} h_2$, iff one of the following holds:
\begin{itemize}
\item $h_1 \models_{i}  r$ and $h_2 \models_{i} r$ and $h_1 =_i h_2$, \\
where $i = \min \{l \; | \; h_1 \models_l r \} = \min \{l \; | \; h_2 \models_l r \}$.
\item $h_1 \models_{irr}  r$ and $h_2 \models_{irr} r$.
\end{itemize}
We say, $h_1$ is at least as preferred as $h_2$ w.r.t. $r$, denoted by $h_1 \succeq_{r} h_2$, iff $h_1 \succ_{r} h_2$ or $h_1 =_{r} h_2$.
\label{def:pref-rule}
\end{definition}
The previous two definitions characterize how probability answer sets are ranked with respect to a boolean combination and with respect to a probability preference rule. Definition (\ref{def:compination}) presents the ranking of probability answer sets with respect to a boolean combination. But, Definition (\ref{def:pref-rule}) presents the ranking of probability answer sets with respect to a probability preference rule. The following definitions specify the ranking of probability answer sets according to a set of probability preference rules.

\begin{definition} [Pareto Preference] Let $\Pi = \langle R_{gen} \cup R_{pref}, \tau \rangle$ be a probability aggregates answer set optimization program and $h_1, h_2$ be probability answer sets of $R_{gen}$. Then, $h_1$ is (Pareto) preferred over $h_2$ w.r.t. $R_{pref}$, denoted by $h_1 \succ_{R_{pref}} h_2$, iff there exists at least one probability preference rule $r \in R_{pref}$ such that $h_1 \succ_{r} h_2$ and for every other rule $r' \in R_{pref}$, $h_1 \succeq_{r'} h_2$. We say, $h_1$ and $h_2$ are equally (Pareto) preferred w.r.t. $R_{pref}$, denoted by $h_1 =_{R_{pref}} h_2$, iff for all $r \in R_{pref}$, $h_1 =_{r} h_2$.
\end{definition}

\begin{definition} [Maximal Preference] Let $\Pi = \langle R_{gen} \cup R_{pref}, \tau \rangle$ be a probability aggregates probability answer set optimization program and $h_1, h_2$ be probability answer sets of $R_{gen}$. Then, $h_1$ is (Maximal) preferred over $h_2$ w.r.t. $R_{pref}$, denoted by $h_1 \succ_{R_{pref}} h_2$, iff
\[
|\{r \in R_{pref} | h_1 \succeq_{r} h_2\}| > |\{r \in R_{pref} | h_2 \succeq_{r} h_1\}|.
\]
We say, $h_1$ and $h_2$ are equally (Maximal) preferred w.r.t. $R_{pref}$, denoted by $h_1 =_{R_{pref}} h_2$, iff
\[
|\{r \in R_{pref} | h_1 \succeq_{r} h_2\}| = | \{r \in R_{pref} | h_2 \succeq_{r} h_1\}|.
\]
\end{definition}
It is worth noting that the Maximal preference definition is more {\em general} than the Pareto preference definition, since the Maximal preference relation {\em subsumes} the Pareto preference relation.

\begin{example} The generator rules, $R_{gen}$, of the probability aggregates probability answer set program, $\Pi = \langle R_{gen} \cup R_{pref}, \tau \rangle$, that represents the two stages stochastic optimization with recourse problem presented in Example (\ref{ex:finance-code}), has $75$ probability answer sets, where the probability answer sets with the least total expected cost are:
\[
\begin{array}{lcl}
I_1 = \{ objective(500,50,200,1330), \ldots \} \\
I_2 = \{ objective(650,0,50,1360), \ldots \} \\
I_3 = \{ objective(550,0,150,1280), \ldots \} \\
I_4 = \{ objective(550,0,200,1340), \ldots \} \\
I_5 = \{ objective(600,0,100,1320), \ldots \} \\
I_6 = \{ objective(600,0,150,1380), \ldots \} \\
I_7 = \{ objective(500,0,200,1240), \ldots \}
\end{array}
\]
The ground instantiation of the probability preference rule in $R_{pref}$ consists of one ground probability preference rule, denoted by $r$, which is
\[
\begin{array}{l}
min_x \{
\\
\qquad \langle 1330 : 1 \; | \; objective(500,50,200,1330) \; \rangle,  \\
\qquad \langle 1360 : 1 \; | \; objective(650,0,50,1360) \; \rangle, \\
\qquad \langle 1280 : 1 \; | \; objective(550,0,150,1280)  \; \rangle, \\
\qquad \langle 1240 : 1 \; | \; objective(500,0,200,1240)  \; \rangle, \\
\qquad \langle 1340 : 1 \; | \; objective(550,0,200,1340) \; \rangle, \\
\qquad \langle 1320 : 1 \; | \; objective(600,0,100,1320) \; \rangle, \\
\qquad \langle 1380 : 1 \; | \; objective(600,0,150,1380) \; \rangle, \\
\ldots \}
\end{array}
\]
Therefore, it can be easily verified that $I_7 \models_1 r$ and
\[
\begin{array}{l}
I_1 \models_{irr} r,
I_2 \models_{irr} r,
I_3 \models_{irr} r,
I_4 \models_{irr} r,
I_5 \models_{irr} r,
I_6 \models_{irr} r
\end{array}
\]
This implies that $I_7$ is the top (Pareto and Maximal) preferred probability answer set and represents the optimal solution for the two stages stochastic optimization with recourse problem described in Example (\ref{ex:finance}). The probability answer set $I_7$ assigns $500$ to $x$, $0$ to $y_1$, and $200$ to $y_2$ with total expected cost $\$1240$, which coincides with the optimal solution of the problem as described in Example (\ref{ex:finance}).
\label{ex:finance-sol}
\end{example}

\section{Properties}

In this section, we show that the probability aggregates probability answer set optimization programs syntax and semantics naturally subsume and generalize the syntax and semantics of classical aggregates classical answer set optimization programs \cite{Saad_ASOG} as well as naturally subsume and generalize the syntax and semantics of classical answer set optimization programs \cite{ASO} under the Pareto preference relation, since there is no notion of Maximal preference relation has been defined for the classical answer set optimization programs.

A classical aggregates classical answer set optimization program, $\Pi^c$, consists of two separate classical logic programs; a classical answer set program, $R^c_{gen}$, and a classical preference program, $R^c_{pref}$ \cite{Saad_ASOG}. The first classical logic program, $R^c_{gen}$, is used to generate the classical answer sets. The second classical logic program, $R^c_{pref}$, defines classical context-dependant preferences that are used to form a preference ordering among the classical answer sets of $R^c_{gen}$.
\\
\\
Any classical aggregates classical answer set optimization program, $\Pi^c = R^c_{gen} \cup R^c_{pref}$, can be represented as a probability aggregates probability answer set optimization program, $\Pi = \langle R_{gen} \cup R_{pref}, \tau \rangle$, where all probability annotations appearing in every probability logic rule in $R_{gen}$ and all probability annotations appearing in every probability preference rule in $R_{pref}$ is equal to $[1,1]$, which means the truth value {\em true}, and $\tau$ is any arbitrary mapping $\tau: {\cal B_L} \rightarrow S_{disj}$. For example, for a classical aggregates classical answer set optimization program, $\Pi^c = R^c_{gen} \cup R^c_{pref}$, that is represented by the probability aggregates probability answer set optimization program, $\Pi = \langle R_{gen} \cup R_{pref}, \tau \rangle$, the classical logic rule
\begin{eqnarray*}
a_1 \; \vee \ldots \vee \; a_k \leftarrow  a_{k+1}, \ldots, a_m, not\; a_{m+1},
\ldots, not\;a_{n}
\end{eqnarray*}
is in $R^c_{gen}$, where $\forall (1 \leq i \leq n)$, $a_i$ is an atom, iff
\[
\begin{array}{r}
a_1:[1,1] \; \vee \ldots \vee \; a_k:[1,1] \leftarrow  a_{k+1}:[1,1], \ldots, a_m:[1,1], \\ not\; a_{m+1}:[1,1],
\ldots, not\;a_{n}:[1,1]
\end{array}
\]
is in $R_{gen}$. It is worth noting that the syntax and semantics of this class of probability answer set programs are the same as the syntax and semantics of the classical answer set programs \cite{Saad_DHPP,Saad_EHPP}. In addition, the classical preference rule
\begin{eqnarray}
C_1 \succ C_2 \succ \ldots \succ C_k \leftarrow l_{k+1},\ldots, l_m, not\; l_{m+1},\ldots, not\;l_{n}
\label{rule:classical-pref}
\end{eqnarray}
is in $R^c_{pref}$, where $l_{k+1}, \ldots, l_{n}$ are literals and classical aggregate atoms and $C_1, C_2, \ldots, C_k$ are boolean combinations over a set of literals, classical aggregate atoms, and classical optimization aggregates iff
\begin{eqnarray}
C_1 \succ C_2 \succ \ldots \succ C_k \leftarrow l_{k+1}:[1,1],\ldots, l_m:[1,1], \notag \\ not\; l_{m+1}:[1,1],\ldots, not\;l_{n}:[1,1]
\label{rule:classical-fuzzy-pref}
\end{eqnarray}
is in $R_{pref}$, where $C_1, C_2, \ldots, C_k$ and $l_{k+1},\ldots, l_n$ in (\ref{rule:classical-fuzzy-pref}) are exactly the same as $C_1, C_2, \ldots, C_k$ and $l_{k+1},\ldots, l_n$ in (\ref{rule:classical-pref}) except that each classical aggregate appearing within a classical aggregate atom or a classical optimization aggregate in (\ref{rule:classical-fuzzy-pref}) involves a conjunction of literals each of which is associated with the probability annotation $[1,1]$, where $[1,1]$ represents the truth value \emph{true}. In addition, any classical answer set optimization program is represented as a probability aggregates probability answer set optimization program by the same way as for classical aggregates classical answer set optimization programs except that classical answer set optimization programs  disallows classical aggregate atoms and classical optimization aggregates.

The following theorem shows that the syntax and semantics of probability aggregates probability answer set optimization programs subsume the syntax and semantics of the classical aggregates classical answer set optimization programs \cite{Saad_ASOG}.

\begin{theorem} Let $\Pi = \langle R_{gen} \cup R_{pref}, \tau \rangle$ be the probability aggregates probability answer set optimization program equivalent to a classical aggregates classical answer set optimization program, $\Pi^c = R^c_{gen} \cup R^c_{pref}$. Then, the preference ordering of the probability answer sets of $R_{gen}$ w.r.t. $R_{pref}$ coincides with the preference ordering of the classical answer sets of $R^c_{gen}$ w.r.t. $R^c_{pref}$ under both Maximal and Pareto preference relations.
\label{thm:theorem}
\end{theorem}
Assuming that \cite{ASO} assigns the lowest rank to the classical answer sets that do not satisfy either the body of a classical preference rule or the body of a classical preference and any of the boolean combinations appearing in the head of the classical preference rule, the following theorems show that the syntax and semantics of the probability aggregates probability answer set optimization programs subsume the syntax and semantics of the classical answer set optimization programs \cite{ASO}.

\begin{theorem} Let $\Pi = \langle R_{gen} \cup R_{pref}, \tau \rangle$ be the probability aggregates probability answer set optimization program equivalent to a classical answer set optimization program, $\Pi^c = R^c_{gen} \cup R^c_{pref}$. Then, the preference ordering of the probability answer sets of $R_{gen}$ w.r.t. $R_{pref}$ coincides with the preference ordering of the classical answer sets of $R^c_{gen}$ w.r.t. $R^c_{pref}$.
\label{thm:1}
\end{theorem}

\begin{theorem} Let $\Pi = \langle R_{gen} \cup R_{pref}, \tau \rangle$ be a probability aggregates probability answer set optimization program equivalent to a classical answer set optimization program, $\Pi^c = R^c_{gen} \cup R^c_{pref}$. A probability answer set $h$ of $R_{gen}$ is Pareto preferred probability answer set w.r.t. $R_{pref}$ iff a classical answer set $I$ of $R^c_{gen}$, equivalent to $h$, is Pareto preferred classical answer set w.r.t. $R^c_{pref}$.
\label{thm:2}
\end{theorem}
Theorem (\ref{thm:theorem}) shows in general probability aggregates probability answer set optimization programs in addition can be used solely for representing and reasoning about multi objectives classical optimization problems by the classical answer set programming framework under both the Maximal and Pareto preference relations, by simply replacing any probability annotation appearing in a probability aggregates probability answer set optimization program by the constant probability annotation $[1,1]$. Furthermore, Theorem (\ref{thm:1}) shows in general that probability aggregates probability answer set optimization programs in addition can be used solely for representing and reasoning about qualitative preferences under the classical answer set programming framework, under both Maximal and Pareto preference relations, by simply replacing any probability annotation appearing in a probability aggregates probability answer set optimization program by the constant probability annotation $[1,1]$. Theorem (\ref{thm:2}) shows the subsumption result of the classical answer set optimization programs.

\section{Conclusions and Related Work}

We developed syntax and semantics of probability aggregates probability answer set optimization programs to represent probability preference relations and rank probability answer sets based on minimization or maximization of some specified criteria to allow the ability to represent and reason and solve probability optimization problems. Probability aggregates probability answer set optimization framework modifies and subsumes the classical aggregates classical answer set optimization presented in \cite{Saad_ASOG} as well as the classical answer set optimization introduced in \cite{ASO}. We shown the application of probability aggregates probability answer set optimization to the two stages stochastic optimization with recourse problem.

To the best of our knowledge, this development is the first to consider a logical framework for representing and reasoning about optimal preferences in general in a quantitative and/or qualitative preferences in answer set programming frameworks. However, qualitative preferences were introduced in classical answer set programming in various forms. In \cite{Schaub-Comp}, preferences are defined among the rules of the logic program, whereas preferences among the literals described by the logic programs are introduced in \cite{Sakama}. Answer set optimization (ASO) \cite{ASO} and logic programs with ordered disjunctions (LPOD) \cite{LPOD} are two answer set programming based preference handling approaches, where context-dependant preferences are defined among the literals specified by the logic programs. Application-dependant preference handling approaches for planning were presented in \cite{Son-Pref,Schaub-Pref07}, where preferences among actions, states, and trajectories are defined, which are based on temporal logic. The major difference between \cite{Son-Pref,Schaub-Pref07} and \cite{ASO,LPOD} is that the former are specifically developed for planning, but the latter are application-independent.

Contrary to the existing approaches for reasoning about preferences in answer set programming, where preference relations are specified among rules and literals in one program, an ASO program consists of two separate programs; an answer set program, $P_{gen}$, and a preference program, $P_{pref}$ \cite{ASO}. The first program, $P_{gen}$, is used to generate the answer sets, the range of possible solutions. The second program, $P_{pref}$, defines context-dependant preferences that are used to form a preference order among the answer sets of $P_{gen}$.

Probability aggregates probability answer set optimization programs distinguish between probability answer set generation, by $R_{gen}$, and probability preference based probability answer set evaluation, by $R_{pref}$, which has several advantages. In particular, $R_{pref}$ can be specified independently from the type of $R_{gen}$, which makes preference elicitation easier and the whole approach more intuitive and easy to use in practice. In addition, more expressive forms of probability preferences can be represented in probability aggregates probability answer set optimization programs, since they allow several forms of boolean combinations in the heads of preference rules.

In \cite{Saad_ASOG}, classical answer set optimization programs have been extended to allow aggregate preferences. This is to allow answer set optimization programs capable of encoding general optimization problems and intuitive encoding of Nash equilibrium strategic games. The classical answer set optimization programs with aggregate preference are built on top of  classical answer set optimization \cite{ASO} and aggregates in classical answer set programming \cite{Recur-aggr}. It has been shown in \cite{Saad_ASOG} that the classical answer set optimization programs with aggregate preferences subsumes the classical answer set optimization programs described in \cite{ASO}.

\bibliographystyle{named}
\bibliography{Saad13LPO}

\end{document}